\documentclass[10pt,twocolumn,letterpaper]{article}
\usepackage{cvpr}
\usepackage[accsupp]{axessibility} 
\usepackage{times}
\usepackage{epsfig}
\usepackage{graphicx}
\usepackage{amsmath}
\usepackage{amssymb}
\usepackage{algorithm}  
\usepackage{algpseudocode}  
\usepackage{booktabs}

\usepackage[pagebackref=true,breaklinks=true,colorlinks,bookmarks=false]{hyperref}

\def\cvprPaperID{2604} 

\begin{document}

\title{PatchFormer: An Efficient Point Transformer with Patch Attention}

\author{Cheng Zhang$^1\thanks{These authors contributed equally.}$, Haocheng Wan$^{1*}$, Xinyi Shen$^2$, Zizhao Wu$^{1}\thanks{Corresponding author: wuzizhao@hdu.edu.cn.}$\\
$^1$Hangzhou Dianzi University, Hangzhou China\\
$^2$University College London, London UK\\
{\tt\small \{zhangcheng828,wuzizhao\}@hdu.edu.cn \{wanhaocheng2022,xinyishen2018\}@163.com}}

\maketitle

\begin{abstract}
The point cloud learning community witnesses a modeling shift from CNNs to Transformers, where pure Transformer architectures have achieved top accuracy on the major learning benchmarks. However, existing point Transformers are computationally expensive since they
need to generate a large attention map, which has quadratic complexity (both in space and time) with respect to input size. To solve this shortcoming, we introduce \textbf{P}atch \textbf{AT}tention (PAT) to adaptively learn a much smaller set of bases upon which the attention maps are computed. By a weighted summation upon these bases, PAT not only captures the global shape context but also achieves linear complexity to input size. In addition, we propose a lightweight \textbf{M}ulti-\textbf{S}cale a\textbf{T}tention (MST) block to build attentions among features of different scales, providing the model with multi-scale features. Equipped with the PAT and MST, we construct our neural architecture called PatchFormer that integrates both modules into a joint framework for point cloud learning. Extensive experiments demonstrate that our network achieves comparable accuracy on general point cloud learning tasks with 9.2$\times$ speed-up than previous point Transformers.
\end{abstract}

\section{Introduction}
Transformer has recently drawn great attention in natural language processing \cite{2018BERT,2017Attention} and 2D vision \cite{DBLP:journals/corr/abs-2010-11929,liu2021swin,wang2021crossformer,vaswani2021scaling} because of its superior capability in capturing long-range dependencies. Self-Attention (SA), the core of Transformer, obtains an attention map by computing the affinities between self $queries$ and self $keys$, generating
a new feature map by weighting the self $values$ with this attention map. Benefitting from SA module, Transformer is capable of modeling the relationship of tokens in a sequence, which is also important to many point cloud learning tasks. 
Hence, plenty of researches have been done to explore Transformer-based point cloud learning architectures.

\begin{figure}
    \centering
    \includegraphics[width=\linewidth]{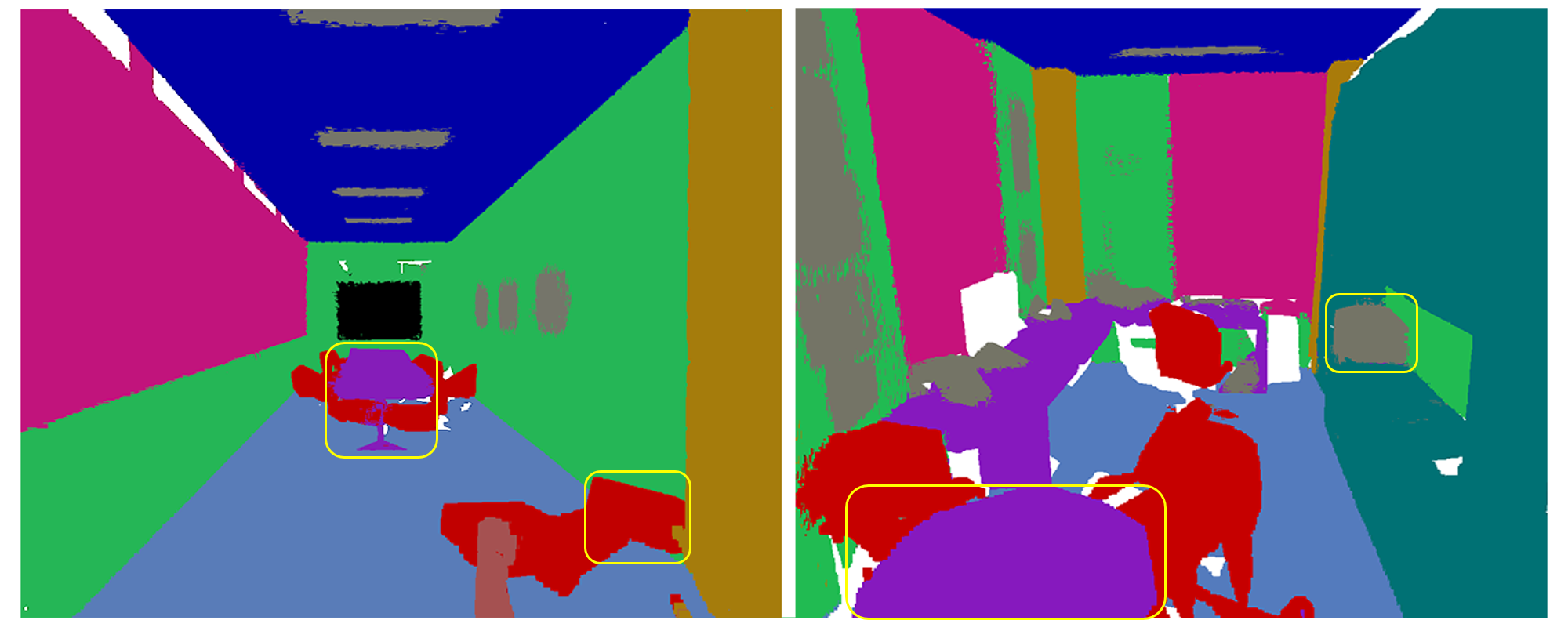}
    \caption{A large indoor scene often consists of small instances (e.g., chair and typewriter) and large objects (e.g., table and blackboard), building the relationships among them requires a multi-scale attention mechanism.}
    \label{fig:multiscale}
\end{figure}
Recently, Nico et al. proposed PT$^1$ \cite{Nico} to extract global features by introducing the standard SA mechanism, which aims to capture spatial point relations and shape information. Guo et al. proposed offset-attention (PCT \cite{guo2020pct}) to calculate the offset difference between the SA features and the input features by element-wise subtraction. Lately, more and more researchers have applied SA module to various point cloud learning tasks and achieved significant performance such as \cite{zhao2020point,cloud}. However, existing point Transformers are computationally expensive because the original SA module needs to generate a large attention map, which has high computational complexity and occupies a huge number of GPU memory. This bottleneck lies in that both the generation of attention map and its usage require the computation with respect to all points. 

Towards this issue, we propose a novel lightweight attention mechanism, namely PAT which calculates the attention map via low-rank approximation \cite{ACFNet,EMANet}. Our key observation is that a 3D shape is composed of its local parts and thus the features of points in the same part should have similar semantics. Based on this observation, we first exploit the intrinsic geometry similarity, cluster local points on a 3D shape as one patch and estimate a base by aggregating the features of all points in the same patch. Then we use a product of self $queries$ and self bases to approximate the global attention map, which can be obtained by computing self $queries$ and self $keys$. 
Notably, the representation of such product is low-rank and discards noisy information from the input.

In addition, to aggregate local neighborhood information, Zhao et al. \cite{zhao2020point} proposed PT$^2$ to build local vector attention in neighborhood point sets, Guo et al. (PCT\cite{guo2020pct}) proposed to use a neighbor embedding strategy to improve point embedding. Though PT$^2$ and PCT have achieved significant progress, there exist problems that restrict their efficiency and performance. First, they wastes a high percentage of the total time on structuring the irregular data, which becomes the efficiency bottleneck \cite{2019Point}. Second, they fails to build the attentions among features of different scales which is very important to 3D visual tasks. As shown in Fig \ref{fig:multiscale}, a large indoor scene often contains small instances (e.g., chair and lamp) and large objects (e.g., table), building the relationships among them required a multi-scale attention mechanism. However, the input sequence of PT$^2$ and PCT is generated from equal-sized points, so only one single scale feature will be preserved in the same layer.
To solve these issues, we present a lightweight \textbf{M}ulti-\textbf{S}cale a\textbf{T}tention (MST) block for point cloud learning, which consists of two steps.
In the first step, our MST block transforms point cloud into voxel grids, sampling boxes with multiple convolution kernels of different scales and then concatenates these grids as one embedding (see Fig \ref{fig:MEB}). Specifically, we propose to use the depth-width convolution (DWConv \cite{dwconv}) on boxes sampling beacuse of both few parameters and FLOPs.
In the second step, we incorporate 3D relative position bias and build attentions to non-overlapping local 3D window, providing our model with strong multi-scale features at a low computational cost.  

Based on these proposed blocks, we construct our neural architecture called PatchFormer (see Fig \ref{fig:netwoek}). Speciﬁcally, we perform the classiﬁcation task on the ModelNet40 and achieve the strong accuracy of 93.5\% (no voting) with 9.2$\times$ faster than previous point Transformers. On ShapeNet and S3DIS datasets, our model also obtains strong performance with 86.5\% and 68.1\% mIoU, respectively.

The main contributions are summarized as following:
\begin{itemize}
\item We present PatchFormer for efficient point cloud learning. Experiments show that our network achieves strong performance with 9.2$\times$ speed-up than prior point Transformers.
\item We propose PAT, the first linear attention mechanism in the point cloud analysis paradigm.
\item We present a lightweight voxel-based MST block, which compensates for previous architectures’ disability of building multi-scale relationship.
\end{itemize}

\section{Related works}


\subsection{Transformer for 2D Vision}
Motivated by the success of Transformers in NLP \cite{2017Attention,relative,yang2019xlnet,Transformer-XL,lee2020biobert}, researchers designed visual Transformers for vision tasks to take advantage of their great attention mechanism. In particular, Vision Transformer (ViT) \cite{DBLP:journals/corr/abs-2010-11929} is the first such example of a Transformer-based approach to match or even surpass convolution neural networks (CNNs) for image classification. Later, Wang et al. \cite{pvt} proposed pyramid structure into transformers, named PVT, greatly decreasing the number of patches in the later layers of the model. Liu et al. \cite{liu2021swin} proposed Swin Transformer whose representation is computed with non-overlapping local windows. Subsequently, Wang et al. \cite{wang2021crossformer} and Chen et al. \cite{crossvit} proposed CrossFormer and CrossViT to study how to learn multi-scale features in Transformers.

Inspired by the cross-scale attention used in CrossFormer and CrossViT for image analysis, we present a voxel-based MST block for point cloud learning that combines voxel grids of different sizes to learn stronger local features. 

\subsection{Point Cloud Learning}
Most existing point cloud learning methods could be classified into two categories in terms of data representations: the \emph{voxel}-based models and the \emph{point}-based models. The \emph{voxel}-based models generally rasterize point clouds onto regular grids and apply 3D convolution for feature learning \cite{zhou2018voxelnet,2016Volumetric,2019VoxSegNet,2018PointGrid,2018SO}. These models are computationally efficient due to their excellent memory locality, but suffer from the inevitable information degrades on the fine-grained localization accuracy \cite{2019PV,2019Point,zhang}. Instead of voxelization, developing a neutral network that consumes directly on point clouds is possible \cite{qi2017pointnet,2019PointWeb,2018Recurrent,2021Semantic,2019Learning,2019Graph,2018Dynamic,2019Linked,graphpbn,RSCNN,CurveNet}. Although these \emph{point}-based models naturally preserve accuracy of point location, they are usually computationally intensive. 

Generally, the voxel-based models have regular data locality and can efficiently encode coarse-grained features, while the point-based networks preserve accuracy of location information and can effectively aggregate fine-grained features. In this paper, we propose PatchFormer to incorporate both the advantages from the two models mentioned above.
\begin{figure*}
    \centering
    \includegraphics[width=\linewidth]{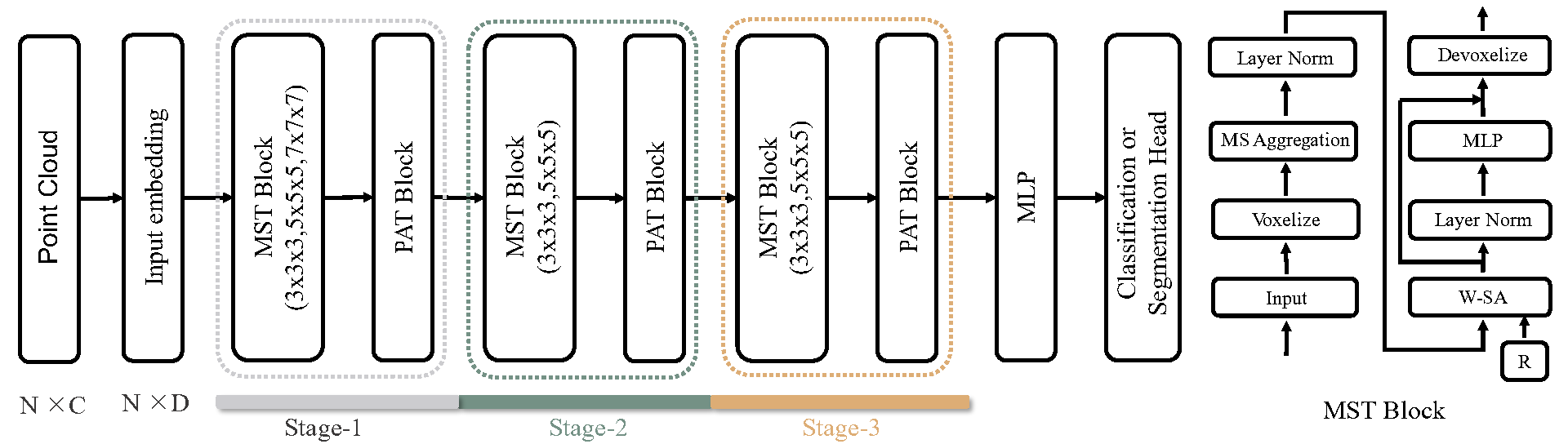}
    \caption{\textbf{The architecture of PatchFormer}: PatchFormer is comprised of three stages and each stage contains two blocks: MST block and PAT bolck. A specialized head (e.g., the classification head) is followed by the final stage for the specific task. \textbf{MST block:} It first voxelizes point cloud into voxel grids, aggregating multi-scale features, and then conducts 3D window-based SA (W-SA) to capture local information. Finally, MST block transforms voxel grids to points and feed them into PAT block. Numbers in MST represent the size of kernels used DWConv, R denotes the relative position bias and W-SA refers to 3D window-based self-attention.}
    \label{fig:netwoek}
\end{figure*}
\subsection{Point Transformers}
Powered by Transformer \cite{2017Attention} and its variants \cite{liu2021swin,DBLP:journals/corr/abs-2010-11929}, the point-based models have recently applied SA to extract features from point clouds and improve performance significantly \cite{guo2020pct,Nico,zhang,PCU,VRPCN,PoinTr}. In particular, PT$^1$ is the first such example of a Transformer-based approach for point cloud learning. Later, Guo et al. and Zhao et al. proposed PCT and PT$^2$ to construct SA networks for general 3D recognition tasks.

However, they suffer from the fact that as the size of the feature map increases, the computing and memory overheads of the original SA increase quadratically. To address this issue, we propose PAT to compute the relation between self $queries$ and a much smaller bases, yet captures the global context of a point cloud as well.

\section{Overview}

An overview of the PatchFormer architecture is presented in Fig \ref{fig:netwoek}. Our method first embeds a point cloud $\mathcal{P}$ into a D dimensional space $F\in \mathbb{R}^{N\times D}$ using a shared MLP, where $N$ is the number of points. We empirically set D = 128, a relatively small value for computational efficiency. Late, we split our model into three stages and each stage is comprised of two blocks: \textbf{M}ulti-\textbf{S}cale a\textbf{T}tention (MST) block and \textbf{P}atch-\textbf{AT}tention (PAT) block. 

As illustrated in Fig \ref{fig:netwoek}, the MST block first voxelizes a point cloud into regular voxel grids and then feeds them into a multi-scale aggregating module. In this module, we sample boxes using three DWConv kernels of different size and concatenate them as one embedding. After that, we limit SA computation to non-overlapping local boxes in order to alleviate the quadratic complexity of the original SA. 
Note that a LayerNorm (LN) layer is applied before W-SA module and MLP module, and a residual connection is applied after each module.
Eventually, we leverage the trilinear interpolation to transform the voxel grids to points.

Like ViT \cite{DBLP:journals/corr/abs-2010-11929}, the PAT block treats each point as a “token” and aggregates global feature by using patch attention. It receives MST block's output as input, estimates a much more compact bases and generates global attention map upon these bases. Note that, our method reduces the complexity (both in space and time) from $\mathcal{O}(N^2)$ of the original SA to $\mathcal{O}(MN)$ ($M<<N$) where $M$ is the number of bases. As a result, the proposed patch attention can conveniently replace the backbone networks in existing point Transformers for various point cloud learning tasks.

Throughout the next sections we use the following notations: the original point cloud with $N$ points is denoted by $\mathcal{P}= \{p_i\}_{i=1}^N \subseteq \mathbb{R}^C $. In the simplest setting of $C$ = 3, each point contains 3D coordinates. $F= \{f_i\}_{i=1}^N \subseteq \mathbb{R}^D $ is the input embedding feature.
\section{Method}
In this section, we first analysis the original SA mechanism, then we detail our novel way to define
attention: patch attention. And finally, we discuss the design of MST block in detail.
\begin{figure*}
    \centering
    \includegraphics[width=\linewidth]{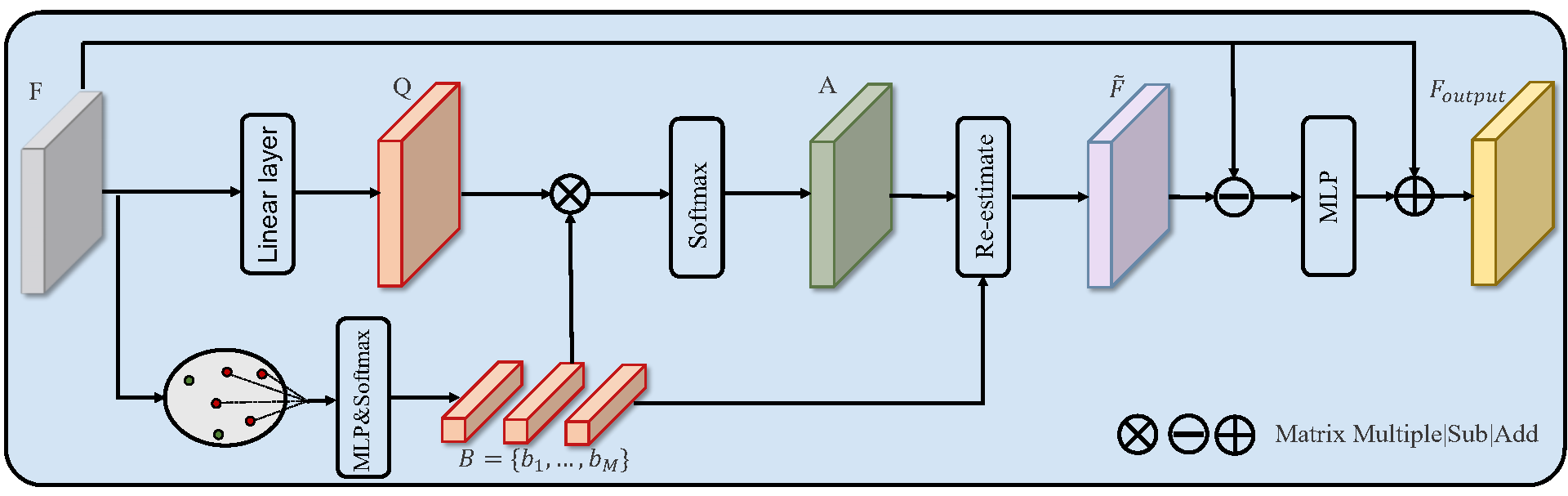}
    \caption{Architecture of PAT block (patch attention). PAT can be seen as an variant to the original self-attention to approximate global map at a lower computational cost.}
    \label{fig:point}
\end{figure*}

\subsection{Self-Attention}
We first revisit the self-attention (SA) mechanism. The standard SA, also called scalar dot-product attention, is a mechanism that calculates semantic affinities among different elements within a sequence of data. Following the terminology in \cite{2017Attention}, let Q, K, V be the $query, key$ and $value$ matrices, respectively, generated by linear transformations of the input features $F \in \mathbb{R}^{N\times D}$ as follows
\begin{alignat}{1}
  (Q,K,V) &= (W_q,W_k,W_v) F,\\
  Q,K,V &\in \mathbb{R}^{N\times D},
\end{alignat}
where $W_q, W_k$ and $W_v$ are the shared learnable linear transformation as illustrated in Fig \ref{fig:point}.

Using the pairwise dot product $QK^{T}\in \mathbb{R}^{N\times N}$, then SA can be formulated as:
\begin{alignat}{1}
    A = (\alpha_{i,j})&=softmax(QK^{T}),\label{map}\\
    F_{out} &= A V,
\label{out}
\end{alignat}
where $A \in \mathbb{R}^{N\times N}$ is the attention map and $\alpha_{i,j}$ is the pair-wise affinity between (similarity of) the $i$-th and $j$-th elements. It is apparent that the output $F_{output}$ is a weighted sum of $V$, where a value gets more weight if the similarity between the keys and values yields a higher attention weighting score.

However, the high computational complexity of $\mathcal{O}(N^2D)$ presents a significant drawback to use of SA. The quadratic complexity in the number of input points makes it infeasible to apply SA to point cloud directly.

\subsection{Patch Attention}
In view of the high computational complexity of the attention mechanism and limitations, we first propose the PAT, which is an augmented version of SA. Unlike prior point Transformers obtain an attention map by computing affinities between self $queries$ and self $keys$, our patch attention (PAT) computes the relation between self $queries$
and a much smaller bases, yet captures the global context of a point cloud.

For simplicity, we consider an input point cloud $\mathcal{P}$ and its corresponding feature map $F$ of size $N\times D$, our proposed PAT is illustrated in Fig \ref{fig:point} which consists of two steps, including \emph{base estimation} and \emph{data re-estimation}.

\textbf{Base Estimation.} In this step, we estimate a compact basis set $B \in \mathbb{R}^{M\times D}$ where $M$ is the number of bases. In particular, we introduce the concept of patch-instance base. For each point cloud $\mathcal{P}$ in the dataset, we over-segment it into $M$ patches ($M<<N$) and based on which, we create $M$ patch-instance bases. In this way, the global shape can be approximated by the set of each patch-instance base, which have a less total number. For simplicity, we use the K-Means algorithm to segment $\mathcal{P}$ into M patches $\{S_1,S_2,...,S_M\}$, M=96, by default in classification task. We define each base as $b_m$ by aggregating the representations of all the points in the $S_m$, it can be described as:
\begin{alignat}{1}
    b_m &= \sum_{f_i\in S_m}w_i(\varphi(f_i)),\\
    B &= \{b_m\}_{m=1}^M \subseteq \mathbb{R}^D.
\label{base}
\end{alignat}
Here, $f_i$ is the representation of point $p_i$, the transformation function $\varphi(\cdot)$ is an MLP with one linear layer and one ReLU nonlinearity, $w_i$ is the normalized degree for $f_i$ belonging to the $S_m$. We use spatial softmax to normalize each patch.

Generally, our base estimation method can adaptively adjust the contribution of all points in the same patch to the base via a data-driven way. Such adaptive adjusting facilitates to fit the intrinsic geometry submanifold.


\textbf{Data Re-estimation.} 
After estimating the bases $B$, we can replace $K$ matrices with $B$ and re-formulate Eq \ref{map} as:
\begin{align}
    A=softmax(QB^T),
\label{gather}
\end{align}
where $A\in \mathbb{R}^{N\times M}$ is the attention map constructed from a compact basis set. After that, the final bases $B$ and attention map $A$ are used to re-estimate the inputs $F$. We formulate a new equation to re-estimate the $F$ using $\tilde{F}$ as follows:
\begin{alignat}{1}
    \tilde{f_i}&= \sum_{m=1}^M A_i^m b_m, \\
    \tilde{F} &= \{\tilde{f_i}\}_{i=1}^N \subseteq \mathbb{R}^D.
\end{alignat}
As $\tilde{F} \in \mathbb{R}^{N\times D}$ is constructed from a compact basis set $B$, it has the low-rank property compared with the input $F$. 

Inspired by PCT \cite{guo2020pct}, we calculate the difference between the estimated features $\tilde{F}$ and the input features $F$ by element-wise subtraction. Finally, we feed the difference into MLP layer and adopt residual connection strategy to help propagate information to higher layers. This step can be formulated as:
\begin{align}
    F_{output} = \phi(\tilde{F}-F)+F,
\end{align}
where $F_{ouput} \in \mathbb{R}^{N\times D}$ is the outputs of our PAT block and $\phi(\cdot)$ is an MLP with one linear layer and one ReLU nonlinearity.

\textbf{Complexity Analysis.}
Compared with the standard SA module, our PAT finds a representative set of bases for points of a point cloud, which reduces the complexity from $\mathcal{O}(N^2)$ to $\mathcal{O}(MN)$ ($M<<N$), where $M$ and $N$ are the number of bases and points, respectively. Moreover, we only need to calculate K-Means algorithm once on the original point cloud $\mathcal{P}$ that can be accelerated in parallel by CUDA. Although the K-Means optimization has an asymptotic complexity $\mathcal{O}(NMC)$, it can be ignored in our network because $M$ is fixed and $C=3$.

\subsection{Multi-Scale Attention}
In this subsection, we detail how our MST block learns multi-scale feature representations in attention models. This block consists of two steps, including \emph{Multi-scale feature aggregating} and \emph{Attention building}.
\begin{figure}
    \centering
    \includegraphics[width=\linewidth]{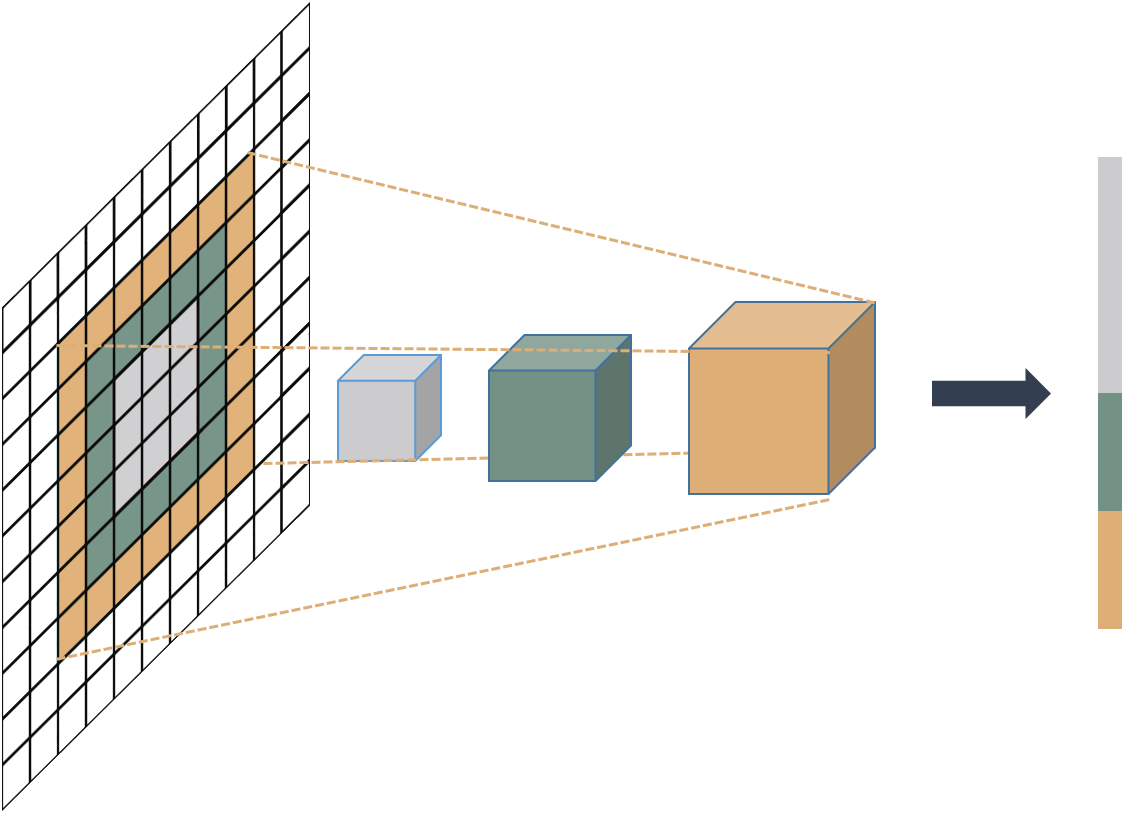}
     \put(-172,65){\footnotesize\bfseries $3\times 3\times 3$}
     \put(-135,60){\footnotesize\bfseries $5\times 5\times 5$}
     \put(-90,55){\footnotesize\bfseries $7\times 7\times 7$}
     \put(-38,48){\footnotesize\bfseries 32 $dims$}
     \put(-38,73){\footnotesize\bfseries 32 $dims$}
     \put(-38,108){\footnotesize\bfseries 64 $dims$}
    \caption{Illustration of multi-scale feature aggregating in MST block on Stage-1. We note that this is a 2D example and can be easily extended to 3D cases. The input voxel grids is sampled by three DWConv kernels (i.e., $3\times 3\times 3$, $5\times 5\times 5$, $7\times 7\times 7$) with stride $1\times 1\times 1$. Each embedding is constructed by projecting and concatenating the three 3D boxes.}
    \label{fig:MEB}
\end{figure}
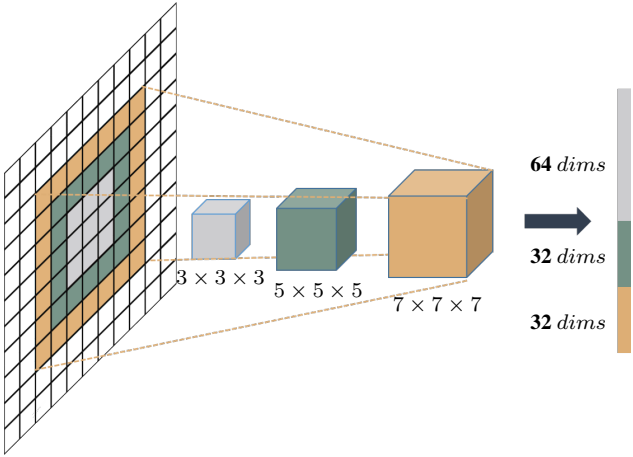

\textbf{Multi-scale Feature Aggregating. }This step is used to generate multi-scale features for each stage. Fig \ref{fig:MEB} illustrates the first MST block, which is ahead of the Stage-1, as an example. we receive voxel grids as input, sampling boxes using three kernels of different size. The strides of three kernels are kept the same so
that they generate the same number of embeddings. As can be seen in Fig \ref{fig:MEB}, every three corresponding boxes own the same center but locate at different scales. These three boxes will be projected and concatenated as one embedding. In practice, the process of sampling and projecting can be implemented through three DWConv layers. Note that we use a lower dimension for large kernels while a higher dimension for small kernels. Fig \ref{fig:MEB} provides the specific allocation rule in its subtable, where a 128 dimensional example is given. Compared with allocating the dimension equally, our scheme reduces computational cost while maintaining the model’s high performance. The MST blocks in other stages work in a similar way. As
shown in Fig \ref{fig:netwoek}, MST blocks in Stage-2/3 use two kernels ($3\times 3\times 3$ and $5\times 5\times 5$). The strides are set as $1\times 1\times 1$. For computing efficiency, DWConv with kernel sizes larger than $5\times 5\times 5$ is implemented by stacking multiple convolutions with kernel size $3\times 3\times 3$ and $5\times 5\times 5$.

\textbf{Attention Building.} To build the attentions among features of different scales. we attempt to conduct the standard SA on multi-scale feature map. However, the computation complexity of the full SA mechanism is quadratic to feature map size. Therefore, it will suffer from huge computation cost for 3D vision tasks that take high resolution feature maps as input, such as semantic segmentation. 

To solve this shortcoming, our MST block limits the SA computation to non-overlapping local 3D windows. In addition, we observe that numerous previous works \cite{2021Video,liu2021swin,rethinking} have shown that it can be advantageous to include relative position bias in SA computation. Thus we introduce 3D relative position bias $R\in \mathbb{R}^{V^3\times V^3}$ as 
\begin{equation}
    F_{output} = softmax(QK^{T}+R)V,
\label{bias}
\end{equation}
where $Q,K,V\in \mathbb{R}^{V^3\times D}$ are the $query$, $key$ and $value$ matrices, and $V^3$  is the number of voxel grids in a local 3D window. Since the relative position along each axis lies in the range $[-V+1, V-1]$, we parameterize a smaller-sized bias matrix $\hat{R}\in \mathbb{R}^{(2V-1)\times(2V-1)\times(2V-1)}$, and values in $R$ are taken from $\hat{R}$.

For cross-window information interaction, existing works \cite{pvt,liu2021swin,vaswani2021scaling} suggested to apply halo or shifted window to enlarge the receptive filed. However, the elements within each Transformer block still has limited attention area and requires stacking more blocks to achieve large receptive filed. In our network, the local attention is building in multi-scale input features. Thus, we don't need to stack more attention layers for cross-window connection or larger receptive filed.

\section{Experiments}
In this section, we evaluate the proposed PatchFormer for different tasks: classiﬁcation, part segmentation, and scene semantic segmentation. Performance is quantitatively evaluated using four metrics: mean class accuracy, overall accuracy (OA), per-class intersection over union (IoU), and mean IoU (mIoU). For fair comparison, we report the measured latency and model size on a RTX 2080 GPU to reﬂect the efﬁciency but evaluate other indicators on a RTX 3090 GPU.

\textbf{Implementation details.} We implement the PatchFormer in PyTorch. We use the SGD optimizer with momentum 0.9 and weight decay 0.0001, respectively. For 3D shape classification on ModelNet40 and 3D object part segmentation on ShapeNetPart, we train for 250 epochs. The initial learning rate is set to 0.01 and is dropped until 0.0001 by using cosine annealing. For semantic segmentation on S3DIS, we train for 120 epochs with initial learning rate 0.5, dropped by 10$\times$ at 50 epochs and 80 epochs.

\subsection{Shape Classification}

\begin{table}  
\centering
\begin{center} 
\begin{tabular}{|l|ccc|}   
\hline
Model  & Input & OA & Latency\\ 
\hline
\multicolumn{4}{|c|}{OA$<$92.5} \\
\hline
PointNet \cite{qi2017pointnet}   & 16$\times$1024 &  89.2  & \textbf{13.6ms}\\ 
PointNet++  \cite{qi2017pointnet++}  & 16$\times$1024 &  91.9  & 35.3ms\\  
SpiderCNN \cite{2018SpiderCNN}   & 8$\times$1024 & 92.4   &82.6ms \\
PointCNN \cite{2018PointCNN}  & 16$\times$1024 & 92.2  & 221.2ms\\  
PointWeb \cite{zhao2019pointweb}  & 16$\times$1024 & 92.3  & $-$\\  
PVCNN \cite{2019Point}  & 16$\times$1024  & 92.4 & 24.2ms\\
\hline
\multicolumn{4}{|c|}{OA$>$92.5} \\
\hline
KPConv \cite{2019KPConv}   &16$\times$6500 & 92.9   &120.5ms \\ 
DGCNN \cite{2018Dynamic}  & 16$\times$1024 & 92.9  &  85.8ms\\
LDGCNN \cite{2019Linked}   & 16$\times$1024 & 92.7  &$-$ \\
PointASNL \cite{2020PointASNL}  & 16$\times$1024 & 93.2  & 923.6ms\\
PT$^1$ \cite{Nico} & 16$\times$1024 & 92.8 & 320.6ms\\
PT$^2$ \cite{zhao2020point}  & 8$\times$1024 & \textbf{93.7}  & 530.2ms \\
PCT \cite{guo2020pct}   & 16$\times$1024 & 93.2 & 92.4ms\\
PatchFormer & 16$\times$1024  & 93.5& \textbf{34.3ms}\\
\hline

\end{tabular}  
\end{center}  
\caption{Results on ModelNet40 \cite{20153D}. Compared with previous Transformer-based models, our PatchFormer achieves the promissing accuracy with 9.2$\times$ measured speed-up on average.} 
\label{cla}
\end{table} 

\textbf{Data.} We evaluate our model on the ModelNet40 \cite{modelnet40} dataset. This dataset contains 12,311 computer-aided design models from 40 man-made object categories, in which 9,843 models are used for training and 2,468 models are used for testing. We follow the experimental conﬁguration of Qi et al. \cite{qi2017pointnet}: (1) we uniformly sample 1,024 points from the mesh faces for each model; (2) the point cloud is re-scaled to ﬁt the unit sphere; and (3) the (x,y,z) coordinates and the normal of the sampled points are used in the experiment. During the training process, randomly scaling, translating and perturbing the objects are adopted as the data augmentation strategy in our experiment.

\textbf{Results.} The results are presented in Table \ref{cla}. The overall accuracy of PatchFormer on ModelNet40 is 93.5\%. It outperforms strong graph-based models such as DGCNN, strong point-based models such as KPConv and excellent attention-based networks such as PointASNL. Remarkably, compared with existing Transformer-based models such as PT$^1$,PT$^2$ and PCT, our model is 9.2$\times$ faster while achieving comparable accuracy.

\subsection{Computational requirements analysis}
\begin{table}  
\centering
\begin{center} 
\begin{tabular}{|p{2.5cm}|cccc|}  
\hline
Model & Params & FLOPs & SDA($\%$) & OA($\%$)\\
\hline
PointNet & 3.47M & 0.45G & 0.0 & 89.2  \\
PointNet++(SSG) & 1.48M & 1.68G &43.5& 90.7\\
PointNet++(MSG) & 1.74M & 4.09G &47.6& 91.9  \\
DGCNN  & 1.81M & 2.43G &57.2& 92.9 \\
\hline
PointASNL  & 3.98M & 5.92G &39.8& 93.1 \\
PT$^1$  & 21.1M & 5.05G &32.5& 92.8\\
PT$^2$  & 9.14M & 17.1G &65.4 & \textbf{93.7}\\
PCT & 2.88M & 2.17G&24.6& 93.2\\
PatchFormer & \textbf{2.45M}& \textbf{1.62G}&\textbf{6.3}&  93.5\\
\hline
\end{tabular}  
\end{center}  
\caption{Computational resource requirements. SDA means the rate of total runtime on structuring the sparse data.} 
\label{computational}
\end{table} 

We now consider the computational requirements of PatchFormer and several other baselines by comparing the floating point operations required (FLOPs) and number of parameters (Params) in Table \ref{computational}. We evaluate these indicators on ModelNet40 dataset. From Table \ref{computational}, we can see that PatchFormer has the lowest memory requirements with only 2.45M parameters and also puts a low load on the processor of only 1.62G FLOPs, yet delivers comparable accurate results of 93.5\%. Notably, we summarize Table \ref{computational} the PatchFormer only spend 6.3\% of the total runtime on structuring the irregular data, which is much lower than previous point Transformers. Compare with its baselines, PatchFormer has not only strong performance but also the lowest computational and memory requirements. These characteristics make PatchFormer suitable for deployment on a edge devices.

\subsection{Object Segmentation}
\begin{figure}
    \centering
    \includegraphics[width=\linewidth]{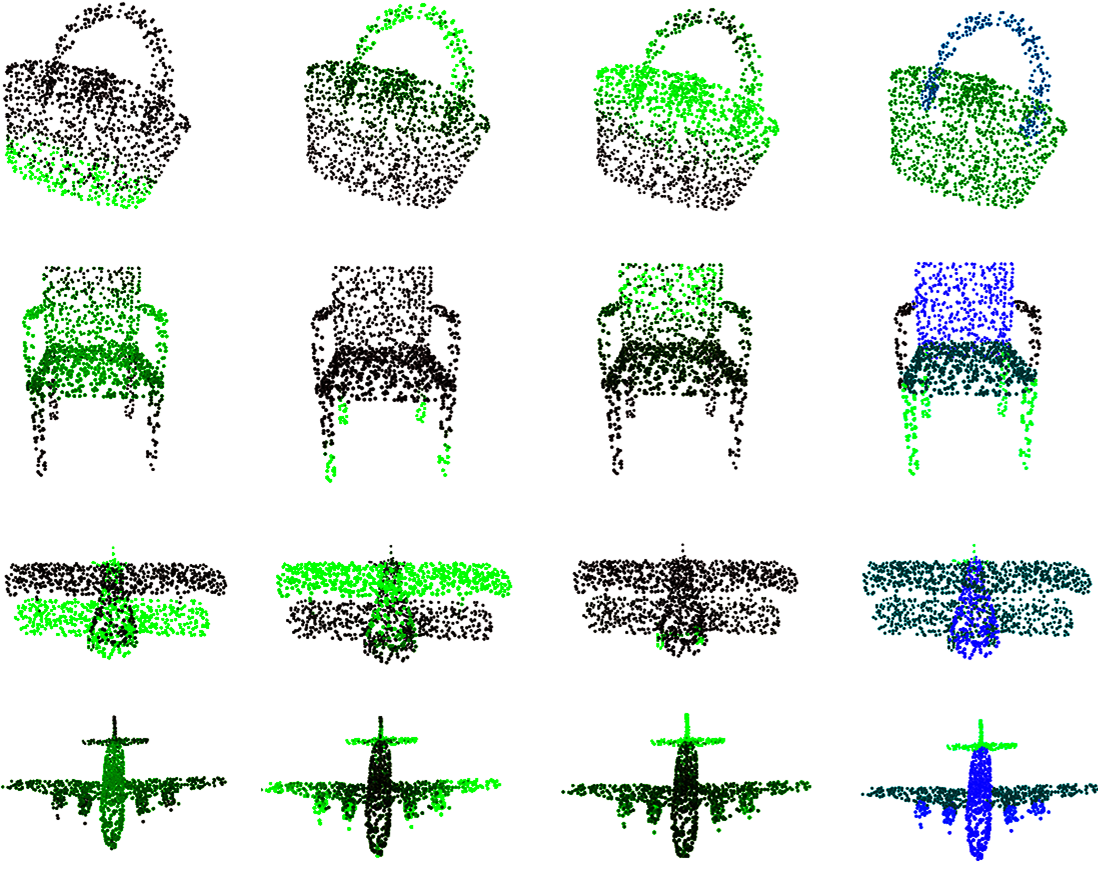}
    \caption{ Attention map and segmentation results on ShapeNet. From left to right: attention maps w.r.t. three selected entries in the bases, segmentation results.}
    \label{fig:atttention}
\end{figure}
\begin{figure}
    \centering
    \includegraphics[width=\linewidth]{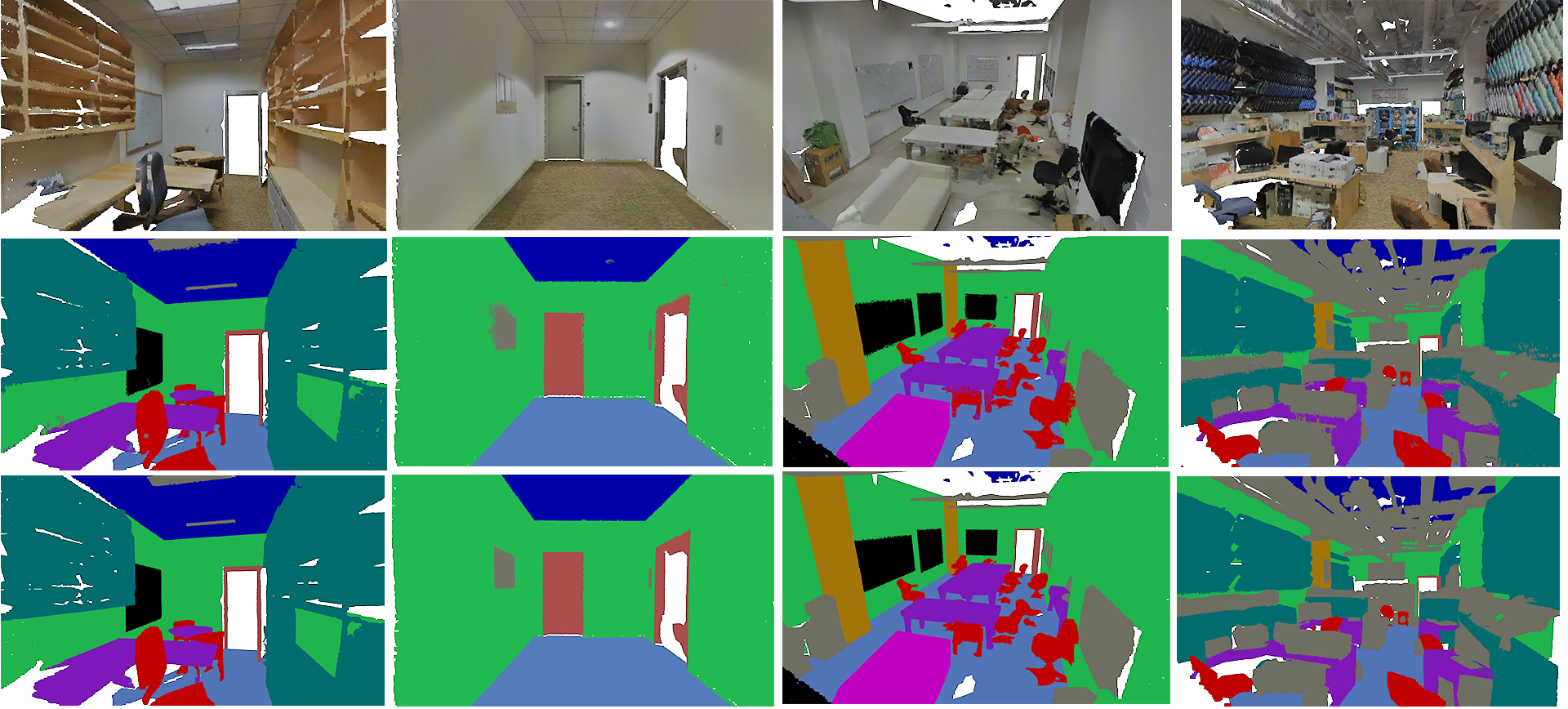}
    \caption{Visualization of semantic segmentation results on the S3DIS dataset. The input is in the top row, PatchFormer prediction is on the middle, the ground truth is on the bottom.}
    \label{fig:sem_vis}
\end{figure}

\begin{table}  
\centering
\begin{center} 
\begin{tabular}{|l|ccc|}   
\hline
Model & Input & mIoU & Latency\\ 
\hline
DGCNN  \cite{2018Dynamic}     & 8$\times$2048 & 85.4 & 96.2ms  \\
PT$^1$ \cite{Nico}     & 6$\times$2048 & 85.9 & 360.4ms  \\
PointCNN \cite{2018PointCNN}   & 8$\times$2048 & 86.1 & 145.6ms  \\
PointASNL \cite{2020PointASNL}   & 6$\times$2048 & 86.1 & 1023.2ms \\
KPConv \cite{2019KPConv}      & 6$\times$6500 & 86.4 & 127.8ms \\
PCT  \cite{guo2020pct}       & 8$\times$2048 & 86.4 & 101.1ms\\
PT$^2$ \cite{zhao2020point}     & 4$\times$2048 & \textbf{86.6} & 560.2ms\\
PatchFormer & 8$\times$2048 & 86.5 & \textbf{45.8ms}\\
\hline
\end{tabular}  
\end{center}  
\caption{Results of part segmentation on ShapeNet Part.} 
\label{part_seg}
\end{table} 

\begin{table}  
\centering
\begin{center} 
\begin{tabular}{|l|ccc|}   
\hline
Model & Input & mIoU & Latency\\ 
\hline
DGCNN       & 8$\times$4096 & 47.1 & 178.1ms  \\
PointCNN \cite{2018PointCNN}   & 4$\times$4096 & 57.3& 282.3ms  \\
PointASNL \cite{2020PointASNL}   & 4$\times$4096 & 62.6 & 1895.2ms \\
PT$^1$ \cite{Nico}     & 4$\times$4096 & 63.1 & 1223.6ms\\
MinkowskiNet \cite{Minkowski}       & $-$ & 65.4 & $-$\\
KPConv  \cite{2019KPConv}    & 4$\times$6500 & 67.1 & 267.5ms \\
PatchFormer & 8$\times$4096 & 68.1 & \textbf{109.8ms}\\
\hline
\end{tabular}  
\end{center}  
\caption{Indoor scene segmentation results on S3DIS, evaluated on Area5. From this table, we can see that PatchFormer outperforms most of previous models in accuracy and efficiency.} 
\label{Area5}
\end{table}

\textbf{Data}. We use the large-scale 3D dataset ShapeNet Parts \cite{shapenet} as the experiment bed. ShapeNet Parts contains 16,880 models (14,006 models are used for training, and 2874 models are used for testing), each of which is annotated with two to six parts, and the entire dataset has 50 different part labels. We sample 2,048 points from each model as input, with a few point sets having six labeled parts. We directly adopt the same train–test split strategy similar to DGCNN \cite{2018Dynamic} in our experiment.

\textbf{Results and Visualization}. From Table \ref{part_seg}, we can see that with similar accuracy, our PatchFormer is 12.4$\times$ faster than PT$^2$ and 2.2$\times$ faster than PCT. Notably, with better accuracy, PatchFormer is 22.7$\times$ faster than PointASNL.
In addition, we randomly select three entity from $B$ in the last layer of our network and show their corresponding attention score of all points. As we can see, each basis corresponds to an abstract concept of the point cloud and the learned attention maps focus on meaningful parts for object segmentation as in Fig \ref{fig:atttention}.

\subsection{Indoor Scene Semantic Segmentation}

\textbf{Data}. We evaluate our model on the S3DIS dataset \cite{DBLP:journals/corr/ArmeniSZS17}, which contains 3D RGB point clouds from six indoor areas of three different buildings. Each point is marked with a semantic label from 13 categories (e.g., board, bookcase, chair, ceiling, and beam) plus clutter. Following a common protocol \cite{2019KPConv,qi2017pointnet}, we divide and sample each room into 1 m $\times$ 1 m blocks, wherein each point is represented by a 9D vector (XYZ, RGB, and normalized spatial coordinates). In addition, the points in each block are sampled into a uniform number of 4,096 points during the training process, and all points are used in the test.

\textbf{Results and Visualization. }The results are presented in Tables \ref{Area5}. From this table we can see that our PatchFormer attains mIoU of 68.1\%, which outperforms graph-based methods such as DGCNN \cite{2018Dynamic}, sparse convolutional networks such as MinkowskiNet \cite{Minkowski}, continuous convolutional networks such as KPConv \cite{2019KPConv}, attention-based models such as PointASNL \cite{2020PointASNL} and point Transformer such as PT$^1$. Remarkably, our PatchFormer also outperforms these powerful model by a large margin in latency.

Fig \ref{fig:sem_vis} shows the PatchFormer’s predictions. We can see that the predictions are very close to the ground truth. PatchFormer captures detailed multi-scale features in complex 3D scenes, which is important in our network.

\subsection{Ablation Studies}
We now conduct a number of controlled experiments that examine specific decisions in the PatchFormer design.

\begin{table}[tb]
\centering
\begin{center}  
\begin{tabular}{|c|ccc|}
\hline
M   & ModelNet40(OA) & ShapeNet(mIoU) & Latency \\
\hline
32  & 91.54           & 84.92          & \textbf{33.25ms}   \\
64  & 92.94           & 85.82          & 33.82ms   \\
96  & \textbf{93.52}  & 86.52          & 34.32ms   \\
128 & 93.50           & \textbf{86.54}          & 35.56ms  \\
\hline
\end{tabular}
\end{center}  
\caption{Ablation study: number of bases M in our network. We report latency on ModelNet40 dataset.}
\end{table}
\textbf{Number of Bases.} We first investigate the setting of the number of bases. The results are shown in Table 5. The best performance of classification task is achieved when $M$ is set to 96. On the one hand, when the bases is smaller ($M$ = 32 or $M$ = 64), the model may not have sufficient context for its predictions. On the other hand, increasing $M$ doesn not give PatchFormer much accuracy benefit but incurs a raise on latency. This also demonstrates the efficiency and effectiveness of our PAT.

\begin{table}[tb]
\centering
\begin{center}  
\begin{tabular}{|c|cc|}
\hline
Ablation  & ModelNet40(OA) & ShapeNet(mIoU)  \\
\hline
w/o MS feature  & 92.85        & 85.22    \\
\hline
MLP  & 92.62           & 85.32                \\
EdgeConv & 93.10          & 85.89          \\
self-attention & 93.29 & 86.22\\
\hline
no rel. pos  & 93.15        & 86.30    \\
\hline
Ours & \textbf{93.52}& \textbf{86.52}\\
\hline
\end{tabular}
\end{center}  
\caption{Ablation study on the multi-scale feature aggregation, PAT and relative bias on two benchmarks. w/o MS feature: all MST block without aggregate multi-scale features. MLP: replace PAT with MLP layer in our architecture. EdgeConv: replace PAT with EdgeConv layer in our architecture. self-attention: replace PAT with self-attention layer in our architecture. rel. pos: the default settings with an additional relative position bias term.}
\label{ablation:MS}
\end{table}

\begin{table}[tb]
\centering
\begin{center}  
\begin{tabular}{|c|cc|}
\hline
Ablation  & ModelNet40(OA) & Latency(ms) \\
\hline
$A^2$ Net \cite{double} & 92.89 & 36.89    \\
EMA Net \cite{EMANet} & 93.02   & 37.27    \\
Linformer \cite{wang2020linformer} & 93.14 & 40.22\\
Performer \cite{performers} & 93.22 & 35.46\\
Ours & \textbf{93.52}& \textbf{34.32}\\
\hline
\end{tabular}
\end{center}  
\caption{We replace our PAT with other linear attention mechanisms. We collect their public code and adapt them to 3D data.}
\label{ablation:attention}
\end{table}

\textbf{Effect of Multi-scale feature aggregating.} We conduct an ablation study on the Multi-scale feature aggregating step. From Table \ref{ablation:MS}, we can see the performance without this step on ModelNet40 and ShapeNet are 92.85\%/85.22\%, in terms of OA/mIoU. It is much lower than the performance with Multi-scale feature (93.52\%/86.52\%). This suggests that the Multi-scale feature is essential in this setting.

\textbf{Impact of PAT.} We investigate the impact of PAT used in the PAT block. From Table \ref{ablation:MS}, we can see that PAT is more effective than the no-attention baseline (MLP). The performance gap between PAT and MLP baseline is significant: 93.52\% vs. 92.62\% and 86.52\% vs. 85.32\%, an improvement of 0.9 and 1.2 absolute percentage points. Compared with EdgeConv baseline, our PAT also achieves improvements of 0.42 and 0.63 absolute percentage points. Notably, our PAT outperforms the self-attention baseline with 0.23 and 0.30 absolute percentage points. We also compare PAT with other linear attention mechanisms in Table \ref{ablation:attention} and find it achieves the best accuracy and running speed. PAT has two obvious advantages. First, we only need to calculate K-Means once on the original point cloud due to the intrinsic geometry similarity, which means that the computational cost of the base estimation can be neglected. Second. PAT based on residual learning is more robust to any rigid transformation of objects.

\textbf{Effect of 3D Relative position bias.} Finally, we investigate the effect of 3D relative position bias used in the MAS block. Table \ref{ablation:MS} shows results. We can see that the PatchFormer with relative position bias yields +0.37\% OA/+0.47\% mIoU on ModelNet40 and ShapeNe in relation to those without position encoding respectively, indicating the effectiveness of the relative position bias.

\section{Conclusion and Future Work}
In this work, we propose a new type of attention mechanism, namely \textbf{P}atch \textbf{AT}tention (PAT) for point cloud learning, which computes a much smaller bases by exploiting the geometric similarity of nearby points.
The reconstructed output of our PAT is low-rank and achieves linear time-space complexity to input size.
Further, we propose a lightweight MST block, building attentions among features of different scales and providing our model with multi-scale features. Based on these modules, we construct PatchFormer for various point cloud learning task. Experiments show that our PatchFormer achieves comparable accuracy and better speed than other point Transformers.

We hope that our work will provide empirical guidelines for new method design and inspire further investigation of the properties of point Transformers. For example, performing K-Means in the points, extracting a patch feature for each cluster, directly reducing the number of tokens in the embedding stage.

\section{Acknowledgements}
This work was partially supported by the Zhejiang Provincial Natural Science Foundation of China (LGF21F20012), the National Natural Science Foundation of China (No.61602139), and the Graduate Scientific Research Foundation of Hangzhou Dianzi University (CXJJ2021082, CXJJ2021083).

{\small
\bibliographystyle{ieee}
\bibliography{egbib}
}

\end{document}